\documentclass[10pt,twocolumn,letterpaper]{article}

\usepackage[pagenumbers]{cvpr} 

\usepackage{graphicx}
\usepackage{amsmath}
\usepackage{amssymb}
\usepackage{booktabs}
\usepackage{multirow}

\usepackage[pagebackref,breaklinks,colorlinks]{hyperref}

\usepackage[capitalize]{cleveref}
\crefname{section}{Sec.}{Secs.}
\Crefname{section}{Section}{Sections}
\Crefname{table}{Table}{Tables}
\crefname{table}{Tab.}{Tabs.}

\def\cvprPaperID{4252} 
\def\confName{CVPR}
\def\confYear{2023}

\begin{document}


\title{CLIP2GAN: Towards Bridging Text with the Latent Space of GANs}

\DeclareRobustCommand*{\IEEEauthorrefmark}[1]{%
    \raisebox{0pt}[0pt][0pt]{\textsuperscript{\footnotesize\ensuremath{#1}}}}
\author{
    Yixuan Wang\IEEEauthorrefmark{1},
    Wengang Zhou\IEEEauthorrefmark{1,3}, 
    Jianmin Bao\IEEEauthorrefmark{2}, 
    Weilun Wang\IEEEauthorrefmark{1},
    Li Li\IEEEauthorrefmark{1},
    Houqiang Li\IEEEauthorrefmark{1,3}
    \\
    \IEEEauthorrefmark{1}CAS Key Laboratory of GIPAS, EEIS Department, University of Science and Technology of China
    \\
    \IEEEauthorrefmark{2}Microsoft Research Asia
    \\
    \IEEEauthorrefmark{3}Institute of Artificial Intelligence, Hefei Comprehensive National Science Center
    \\
    \href{mailto:wyx2017@mail.ustc.edu.cn,wwlustc@mail.ustc.edu.cn}{\{wyx2017, wwlustc\}@mail.ustc.edu.cn}, 
    \href{mailto:jianbao@microsoft.com}{jianbao@microsoft.com}, 
    \href{mailto:zhwg@ustc.edu.cn,lil1@ustc.edu.cn,lihqc@ustc.edu.cn}{\{zhwg,lil1,lihq\}@ustc.edu.cn}
}


\twocolumn[{%
\renewcommand\twocolumn[1][]{#1}%

\maketitle

\begin{center}
    \centering
    \includegraphics[width=0.98\linewidth]{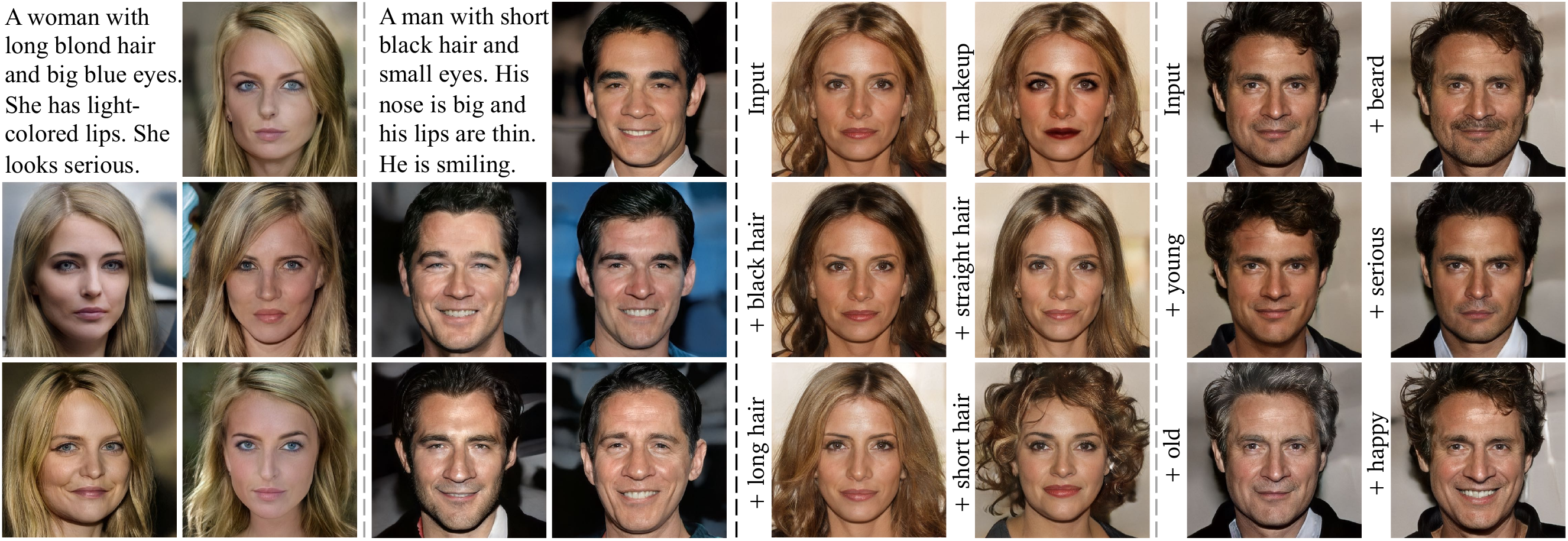}
    \vspace{-0.3cm}
    \captionof{figure}{Text-guided image generation and editing results of our proposed CLIP2GAN. The left shows diverse generation results given a text description. The right shows the result of image editing using text. The images generated by our framework are realistic and accurate.}
    \label{fig:pic1}
    \vspace{-0.1cm}
\end{center}%
}]


\begin{abstract}
\vspace{-0.45cm}

In this work, we are dedicated to text-guided image generation and propose a novel framework, \emph{i.e.}, CLIP2GAN, by leveraging CLIP model and StyleGAN. The key idea of our CLIP2GAN is to bridge the output feature embedding space of CLIP and the input latent space of StyleGAN, which is realized by introducing a mapping network. In the training stage, we encode an image with CLIP and map the output feature to a latent code, which is further used to reconstruct the image. In this way, the mapping network is optimized in a self-supervised learning way. In the inference stage, since CLIP can embed both image and text into a shared feature embedding space, we replace CLIP image encoder in the training architecture with CLIP text encoder, while keeping the following mapping network as well as StyleGAN model. As a result, we can flexibly input a text description to generate an image. Moreover, by simply adding mapped text features of an attribute to a mapped CLIP image feature, we can effectively edit the attribute to the image. Extensive experiments demonstrate the superior performance of our proposed CLIP2GAN compared to previous methods.

\end{abstract}

\vspace{-0.5cm}
\section{Introduction}
\label{sec:intro}

\vspace{-0.1cm}
In recent years, generative models based on  GAN~\cite{goodfellow2014generative} have achieved remarkable success in various tasks including image-to-image translation~\cite{richardson2021encoding, huang2021unsupervised}, image inpainting~\cite{abdal2020image2stylegan++, cheng2022inout}, video generation~\cite{tulyakov2018mocogan}, \emph{etc.}
Specifically, in image generation, the quality of images generated by GANs has been improving with the emergence of several advanced GANs~\cite{karras2018progressive, karras2019style, karras2020analyzing, karras2021alias} that are capable of generating high-resolution and high-fidelity images.
Among them, StyleGAN~\cite{karras2019style} disentangles image attributes in the intermediate latent space, which allows for various image editing and manipulation tasks\cite{abdal2019image2stylegan, bau2019inverting, zhu2020domain, shen2020interfacegan, shen2020interpreting, collins2020editing, wu2021stylespace}.
Previous work tends to achieve image applications by controlling the latent code features directly, which makes it difficult to involve explicit intentions.
Since text can express what people need precisely, it is natural to explore whether text can be utilized to control the latent space directly to achieve text-guided image generation and image editing tasks.
In this paper, we propose a novel framework, \emph{i.e.}, CLIP2GAN, for text-guided image generation and editing. 
Technically, our task can be decomposed into two subtasks. 
First, the input text is transferred to an embedding feature space. 
Second, the text feature is used to generate an image. 
For each subtask, there are successful solutions in literature, such as CLIP model~\cite{radford2021learning} for the first subtask and StyleGAN~\cite{karras2019style} for the second subtask. 
However, CLIP and StyleGAN are decoupled since the feature from CLIP is not aligned with the latent code of StyleGAN. 
To bridge CLIP and StyleGAN, we introduce a mapping network, which transfers the feature embedding of CLIP to the latent space of StyleGAN.
Specifically, in the training stage, we encode an image with CLIP and map the output feature to a latent code, which is further used to reconstruct the image. 
In this way, the mapping network is optimized in a self-supervised learning way. 
In the inference stage, since CLIP can embed both image and text into a shared space, we replace CLIP image encoder in the training architecture with CLIP text encoder. 
Consequently, we can flexibly input a text description to generate a face image. 

In CLIP model, an image is encoded into a 512-dimensional feature, which inevitably loses some detailed visual clue and results in the generated images missing information such as hair details, skin texture, background, \emph{etc.}
To improve the quality of generated images, we introduce an additional discriminator trained adversarially with the mapping network, thus ensuring that the mapping network can generate as realistic and high-resolution images as possible.
Besides, we also add noise to the CLIP image feature.
On the one hand, it further supplements the details lost by CLIP.
On the other hand, inspired by the mode seeking regularization~\cite{mao2019mode}, we maximize the ratio of the distance between the output images after adding noise to the distance between their corresponding latent codes, so that the generated images are diverse while maintaining high fidelity.
Unlike previous methods~\cite{zhang2017stackgan, zhang2018stackgan++, xu2018attngan, xia2021tedigan, zhang2021cross, tao2022df} using image-text pairs for training, our framework generates high-quality face images in a zero-shot way, which means we use image data as input for training and text data as input for testing.
Thanks to CLIP's multi-modal embedding space, our text-free training approach is not constrained by large-scale image-text datasets that require precise manual annotation and are not easily available in practice~\cite{xu2018attngan, ding2021cogview, ramesh2021zero, zhang2021cross, zhu2019dm}, and can generate text-guided realistic and accurate images.
Compared with those who use image-text pairs for training for text-guided image generation, our model is implemented in a minimal-cost text-free training approach, but still maintains extremely high fidelity and quality.

Besides image generation, we explore CLIP2GAN on image editing task, where real images can be edited directly using textual attributes to adjust their expressions, hair characteristics, age, \emph{etc.}
Taking advantage of StyleGAN, we locate different feature orientations of different text descriptions in the latent space.
By imposing different orientations in the latent code of the image, the semantics of the image is controlled and changed in a fine-grained manner with high quality.
Specifically, CLIP2GAN is convenient and flexible for image editing, because instead of optimizing the model for specific text like StyleCLIP~\cite{patashnik2021styleclip}, we can simply use arithmetic operations to add attributes to images by adding mapped CLIP features of text describing the attributes to mapped CLIP features of images.

To evaluate our proposed method, we perform experiments on the CelebA-HQ~\cite{karras2018progressive} dataset.
Both quantitative and qualitative results validate that compared with previous methods, our text-free training model can generate high-fidelity and diverse results and realize image manipulation, achieving superior performance over most existing models trained using full image-text pairs.
Some example results are shown in \cref{fig:pic1}.

In summary, our contributions are as follows:
\begin{itemize}
    \vspace{-2mm}
    \item We propose a new framework, \emph{i.e.}, CLIP2GAN, that enables text-guided generation tasks with text-free training, generating diverse and high-quality images given the same input text.
    \vspace{-2mm}
    \item We apply our framework to image manipulations that allows the editing of real images directly with textual attributes.
    \vspace{-2mm}
    \item Extensive experiments on public datasets demonstrate the validity and superiority of our framework. The generated images of our method have higher evaluation quality and better visual performance.
\end{itemize}
\section{Related Work}

\begin{figure*}
    \centering
    \includegraphics[width=\linewidth]{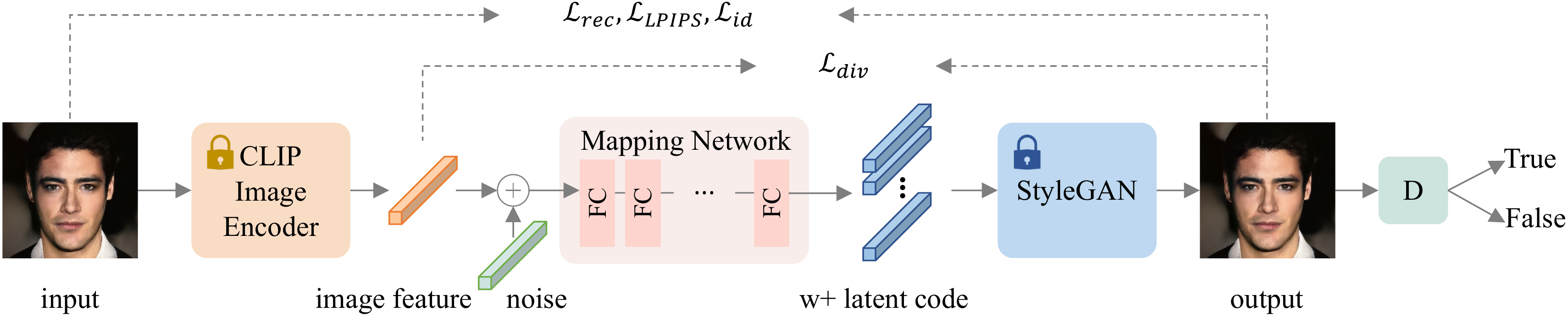}
    \vspace{-0.6cm}
    \caption{Architecture of CLIP2GAN for training. CLIP2GAN takes the face image as input into the CLIP image encoder. The 12-layer mapping network is learned by inverting CLIP image features back into their original inputs using pre-trained StyleGAN. We add the discriminator, \emph{i.e.}, D, and noise respectively to complement the details lost by CLIP and to encourage the diversity of the generated images.}
    \label{fig:pic2}
    \vspace{-0.55cm}
\end{figure*}

\vspace{-0.05cm}
\subsection{High-quality Face Generation}
\vspace{-0.05cm}

Due to the great potential of GAN in generating realistic and high-resolution images, it is widely used in image generation and applications~\cite{lample2017fader, ledig2017photo, yu2019free, wang2018video}.
In particular, high-quality face generation has been an attractive problem in image generation.
PGGAN~\cite{karras2018progressive} first proposes the idea of resolution progressive generation to generate high-definition face images, which first discovers large-scale structures and then focuses on fine details.
StyleGAN~\cite{karras2019style, karras2020analyzing, karras2021alias} introduces a novel style-based generator architecture that generates face images with high fidelity and high resolution.
It controls the visual features represented in each layer individually, which can be coarse features influenced by style (\emph{e.g.}, pose, face shape, identity features, \emph{etc.}) or detailed features influenced by noise (\emph{e.g.}, pupil, hair, wrinkles, \emph{etc.}).
Unlike subsequent studies~\cite{karras2020training, afifi2021histogan, alaluf2022hyperstyle} that mostly introduces different mechanisms or structures on StyleGAN, we use pre-trained StyleGAN to achieve high-quality face generation at a low cost and high efficiency.

\vspace{-0.1cm}
\subsection{Joint Vision-language Models}
\vspace{-0.1cm}


With the remarkable progress in both computer vision and natural language processing, researchers turn their attentions to joint vision-language (VL)  models for many task-specific VL problems, including image captioning~\cite{xu2015show, karpathy2014deep, vinyals2015show}, visual question and answer (VQA)~\cite{antol2015vqa, yang2016stacked}, image text matching~\cite{karpathy2014deep, huang2017instance}, \emph{etc.}, which are tailored for specific problems and each model only solves one task.
After the introduction of the transformer~\cite{vaswani2017attention}, BERT~\cite{devlin2018bert} has achieved unprecedented success in various language tasks.
A recent development, CLIP~\cite{radford2021learning}, pre-trained using over 400 million image-text pairs based on contrastive learning, learns a multi-modal co-embedding space and estimates the semantic similarity between texts and images.
The robustness of learned joint representation enables CLIP to offer high performance and excellent generalization on various tasks.

\vspace{-0.05cm}
\subsection{Text-guided Generation and Manipulation}
\vspace{-0.05cm}

Text-guided image generation is an interesting topic in image generation, where GAN-based models show better sample quality.
StackGAN~\cite{zhang2017stackgan, zhang2018stackgan++} stacks several generators and discriminators to improve the resolution of generated images in multiple stages.
AttnGAN~\cite{xu2018attngan} introduces a cross-modal attention mechanism to explore fine-grained text and image representations.
XMC-GAN~\cite{zhang2021cross} utilizes contrastive learning for image generation.
TediGAN~\cite{xia2021tedigan} trains an encoder to map the text into the latent space of StyleGAN~\cite{karras2019style}.
DF-GAN~\cite{tao2022df} proposes a one-stage backbone for the direct synthesis of high-resolution images.
Compared to most previous work, we have achieved better performance without using text training.

Similar to text-guided generation, manipulating a given image using text produces results containing the desired properties. 
The difference is that the edited result should change the parts related to the text and retain the rest of it.
For instance, Dong \emph{et al.}~\cite{dong2017semantic} propose an encoder-decoder structure for text-guided manipulation, and Li \emph{et al.}~\cite{li2020manigan} generate high-quality images through a multi-stage network.
Unlike most text-guided image manipulation based on a multi-stage framework, we propose a unified framework that allows text-guided image generation and image manipulation without requiring multi-stage processing.

\vspace{-0.1cm}
\section{Method}
\vspace{-0.1cm}

In this paper, we propose a novel framework, CLIP2GAN, for text-guided image generation without text training (see \cref{fig:pic2}).
Our framework is capable of generating accurate and high-quality images under fine-grained text control without training on paired image-text data.
Benefiting from the diversity loss we designed, the multi-modal generation of face images that matches a specific text description is also supported.
Furthermore, we explore our framework for image manipulation tasks, where real images are edited using text to adjust their attributes, \emph{e.g.}, expressions, hair color, and age, with high fidelity and reliability.
The rest of this section is organized as follows. 
We first introduce the overall structure of CLIP2GAN.
Then the loss functions utilized in our framework are discussed.
Finally, we present the image manipulation application implemented using our framework.
\vspace{-0.05cm}

\subsection{CLIP2GAN}
\vspace{-0.05cm}

\cref{fig:pic2} gives an overview of our framework, which consists of a pre-trained vision-language model (CLIP), a mapping network, and a pre-trained generation model (StyleGAN).
Unlike previous work that uses a large number of image-text pairs for model training, our approach achieves text-free training by establishing a mapping relationship between the CLIP multi-modal embedding space and the StyleGAN latent space.


On the one hand, to achieve text-guided image generation without text training, we generate pseudo-text features by leveraging the image-text feature alignment of a pre-trained model.
We require a universal multi-modal embedding space where the paired text and image features can be well aligned.
The recent vision-language model CLIP achieves this by pre-training a large number of image-text pairs through Contrastive Learning, which is exactly what we need.
On the other hand, given that StyleGAN has excellent latent space, we can perform a series of manipulations on the generated images by changing its latent code.
We take advantage of the pre-trained StyleGAN2 as the model for image generation.
With the help of StyleGAN's latent space, we can get high-quality text-guided image generation and image editing of images.

To generate images from text, we build a bridge between CLIP and StyleGAN through a mapping network. 
With this mapping network, it is possible to obtain feature representations of text or images in the latent space of StyleGAN and thus generate images using StyleGAN. 
The source image $x$ is taken as the input of CLIP and the image encoder of CLIP is used to obtain the image features $f_{img}$, \emph{i.e.} pseudo-text features $\tilde{f}_{text}$, in the multi-modal embedding space of CLIP. 
It is formulated as follows,
\vspace{-0.1cm}
\begin{equation}
    \tilde{f}_{text} = f_{img} = C_{img}(x),
    \label{eq:clipfimg}
    \vspace{-0.1cm}
\end{equation}
where $C_{img}(\cdot)$ denotes the image encoder of the CLIP model~\cite{radford2021learning}.
The image features $f_{img}$ are mapped to latent codes $z$ of StyleGAN in $w+$ space by the mapping network as the input of the pre-trained StyleGAN, and the image $x'$ is generated by StyleGAN. 
$x'$ is expressed as follows,
\vspace{-0.1cm}
\begin{equation}
    x' = G(M(C_{img}(x))),
    \label{eq:CLIP2GAN}
    \vspace{-0.1cm}
\end{equation}
where $M(\cdot)$ denotes the mapping network and $G(\cdot)$ denotes the pre-trained StyleGAN model.
By learning the consistency of the source image $x$ and the generated image $x'$, our generative model is implemented.

The mapping network is capable of mapping the multi-modal embedding space of CLIP into the $w+$ latent space of StyleGAN, which makes it possible to invert CLIP features back into the source images using StyleGAN.
Our proposed mapping network is a 12-layer fully connected layer with each layer post-connected to the activation layer Leaky ReLU~\cite{maas2013rectifier}.
The mapping network converts a 512-dimensional text feature $f_{text}$ or image feature $f_{img}$ from the multi-modal embedding space of CLIP into the $w+$ space of StyleGAN for obtaining an 18$\times$512-dimensional latent code $\mathbf{z}$, which is formulated as follows,
\vspace{-0.1cm}
\begin{equation}
    \mathbf{f}^{i+1} = g(\mathbf{w}_{i} \cdot \mathbf{f}^{i} + \mathbf{b}_{i}),\ \mathbf{f}^{0} = f_{text}\ or\ f_{img},
    \label{eq:mappinglayers}
    \vspace{-0.1cm}
\end{equation}
where $\mathbf{f}^{i+1}, \mathbf{f}^{i}\in\mathbb{R}^{H\times W\times C}$ are the input and output features, respectively. $g(\cdot)$ denotes the Leaky ReLU activation.

\begin{figure}
    \centering
    \includegraphics[width=\linewidth]{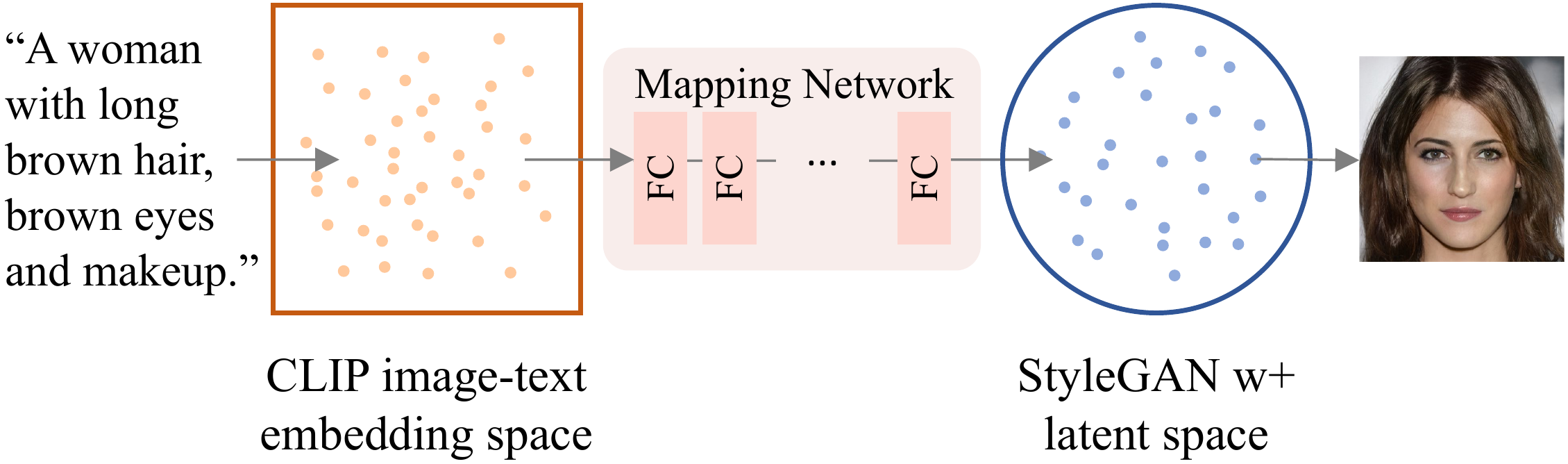}
    \vspace{-0.7cm}
    \caption{Schematic of CLIP2GAN for testing. We use the trained mapping network as shown in \cref{fig:pic2} to bridge the image-text embedding space of CLIP and the $w+$ latent space of StyleGAN and take the text as input to generate an image corresponding to the text description.}
    \label{fig:pic2_2}
    \vspace{-0.6cm}
\end{figure}

However, we discover that when using only a simple mapping network, CLIP suffers from missing details in both the inversion of the CLIP image features and generated images.
Although it can ensure that the major features are presented, the image details, especially the hair and skin textures, backgrounds, \emph{etc.}, are lost to varying degrees, which is attributed to the fact that CLIP only extracts the major 512-dimensional features of the image and ignores the others.
To tackle this issue, we introduce a discriminator to determine the truthfulness of the obtained images. 
It is trained adversarially with the mapping network to complement the image details and improve the generation quality without affecting the feature representations.


Furthermore, when constraints are applied only between the source images and the reconstructed generated images, it is observed that the generated image examples lack diversity.
This is because the loss between pixels mainly focuses on the reconstructed identical images rather than the diverse images, although diversity contributes to the performance.
To this end, we apply diversity loss on the latent space to generate diverse images as shown in \cref{fig:pic2}.
Inspired by the mode seeking regularization~\cite{mao2019mode}, the diversity loss is achieved by maximizing the ratio of the distance between the output images to the distance between the corresponding latent codes, 
and it encourages the generation of distinctive results when different noise vectors are brought in.
Inputting an arbitrary specific text description, we are able to generate multiple images that match the text features but are unique with an additional Gaussian noise of $\mu=1,\sigma^2=0.36$.

Considering the image-text alignment property of CLIP's multi-modal embedding space, we input the source image and learn the mapping of CLIP image features, \emph{i.e.}, pseudo-text features, to the latent space by image reconstruction.
That means training using image data, as shown in \cref{fig:pic2}.
Meanwhile, during testing, the target text description is input and text-guided image generation is performed utilizing CLIP text features, which is shown in \cref{fig:pic2_2}.
Through the above process, a solution for text-guided image generation without text training is achieved.

\subsection{Loss Functions}

We utilize several objective functions, \emph{i.e.}, reconstruction loss, perceptual loss, adversarial loss, identity loss, and diversity loss to optimize our framework.

\noindent \textbf{Reconstruction loss.}
For a given image $x$, the reconstructed image $x'$ is generated by our framework. 
We develop a reconstruction loss to guarantee the pixel alignment between $x$ and $x'$. The reconstruction loss is formulated as follows,
\begin{equation}
    \mathcal{L}_\text{rec} = \|x - x'\|_2,
    \label{eq:l2loss}
\end{equation}
where $\|\cdot\|_2$ denotes the $l_2$ distance.

\noindent \textbf{Perceptual loss.}
The reconstruction loss using $l_2$ distance assumes that the data fit a Gaussian distribution, which leads to producing smoother images.
We introduce a perceptual loss, \emph{i.e.}, LPIPS loss~\cite{zhang2018unreasonable} to measure the difference between the source image $x$ and the generated image $x'$ to improve the smoothing problem caused by the reconstruction loss.
LPIPS learns to reconstruct the reverse mapping of $x$ from $x'$ and prioritizes their perceptual similarity, which is formulated as follows,
\begin{equation}
    \mathcal{L}_\text{LPIPS} = \sum_{l}\frac{1}{H_{l}W_{l}}\sum_{h,w}\|w_{l}\odot(y_{hw}^{l}-y'^{l}_{hw})\|_{2}^{2},
    \label{eq:lpipsloss}
\end{equation}
where $y^{l}$ and $y'^{l}$ are the feature extracted from the $l$-th layer of a pre-trained AlexNet~\cite{krizhevsky2017imagenet}.

\noindent \textbf{Adversarial loss.}
The discriminator is trained adversarially with the mapping network which is considered as a generator.
For generator ${G(\cdot)}$ and discriminator ${D(\cdot)}$, we use the WGAN-GP losses~\cite{arjovsky2017wasserstein, gulrajani2017improved}, which are formulated as follows,
\begin{equation}
    \mathcal{L}_{G} = \mathrm{E}_{x'}[\log(1-D(x')],
    \label{eq:gganloss}
\end{equation}
\begin{equation}
  \begin{aligned}
      \mathcal{L}_{D} = &- \{\mathrm{E}_{x}[\log D(x)] + \mathrm{E}_{x'}[\log(1-D(x'))]\} \\
      &+ \lambda\{\mathrm{E}_{x'}[\|\nabla_{x'}D(x')\|_{2}-1]^2\},
  \end{aligned}
  \label{eq:dganloss}
\end{equation}
where $x$ and $x'$ denote the source image and reconstructed image, respectively.

\noindent \textbf{Identity loss.}
To ensure that the identity features of the faces are unchanged, we present an identity loss. 
Using an effective arcface model~\cite{deng2019arcface}, our identity loss calculates the similarity of the identity features of the faces of $x$ and $x'$, which is formulated as follows,
\begin{equation}
    \mathcal{L}_\text{id} = 1 - \frac{f_\text{arc}(x)\cdot f_\text{arc}(x')}{[f_\text{arc}(x)\cdot f_\text{arc}(x)][f_\text{arc}(x')\cdot f_\text{arc}(x')]},
    \label{eq:idloss}
\end{equation}
where $f_\text{arc}(\cdot)$ denotes an arcface classifier.

\noindent \textbf{Diversity loss.}
As shown in \cref{fig:pic2}, in the multi-modal embedding space of CLIP the standard normally distributed noise of $\mu=0,\sigma^2=1$ is added to the obtained CLIP image feature $f_{img}$ to derive the image feature $f_{img_1}$.
The reconstructed images $x',x_1'$ are generated respectively, and then the diversity loss is formulated as follows,
\begin{equation}
    \mathcal{L}_\text{div} = \frac{d_{f}(f_{img},f_{img_1})}{d_{I}(x',x_1')},
    \label{eq:msloss}
\end{equation}
where $d_*(\cdot)$ denotes the distance calculation and we use $l_1$ distance. 

\noindent \textbf{Overall loss.}
The overall loss for CLIP2GAN is the weighted summation of the above losses, which is formulated as follows,
\vspace{-0.1cm}
\begin{equation}
    \begin{aligned}
        \min_{M}\mathcal{L}_{M} = &\mathcal{L}_\text{rec} + \lambda_\text{LPIPS}\mathcal{L}_\text{LPIPS} + \lambda_{G}\mathcal{L}_{G} \\
        \vspace{-0.1cm}
        &+ \lambda_\text{id}\mathcal{L}_\text{id} + \lambda_\text{div}\mathcal{L}_\text{div},
    \end{aligned}
    \vspace{-0.1cm}
    \label{eq:loss}
\end{equation}
where $\lambda_\text{LPIPS}, \lambda_{G}, \lambda_\text{id}$ and $\lambda_\text{div}$ are the trade-off parameters balancing different losses.

\begin{figure}
    \centering
    \includegraphics[width=\linewidth]{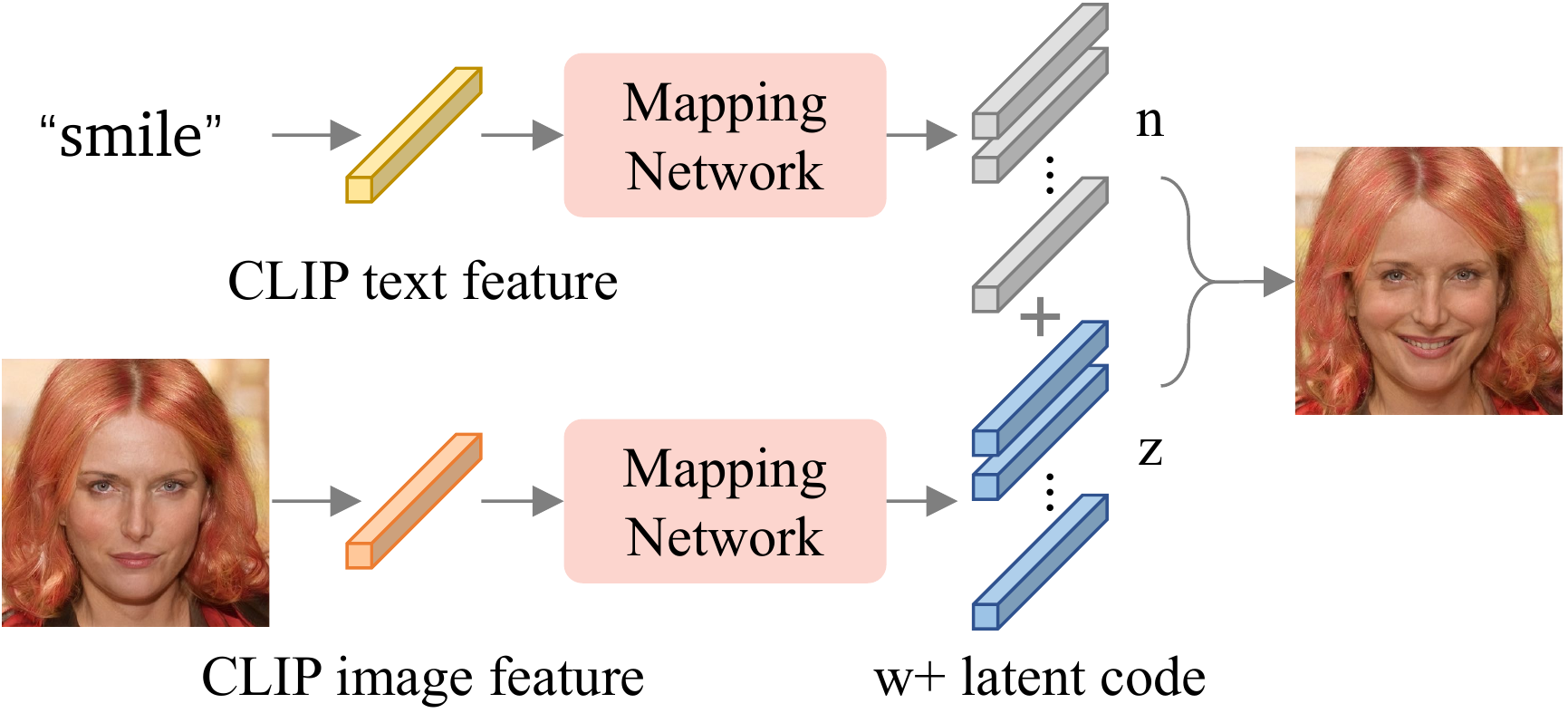}
    \vspace{-0.6cm}
    \caption{Text-guided image editing using CLIP2GAN. The semantic direction $\mathbf{n}$ is discovered in the $w+$ latent space of StyleGAN using our model, which modifies the latent code $\mathbf{z}$ of the image and thus the edited image is generated.}
    \label{fig:pic3}
    \vspace{-0.55cm}
\end{figure}
\begin{figure*}
    \centering
    \includegraphics[width=\linewidth]{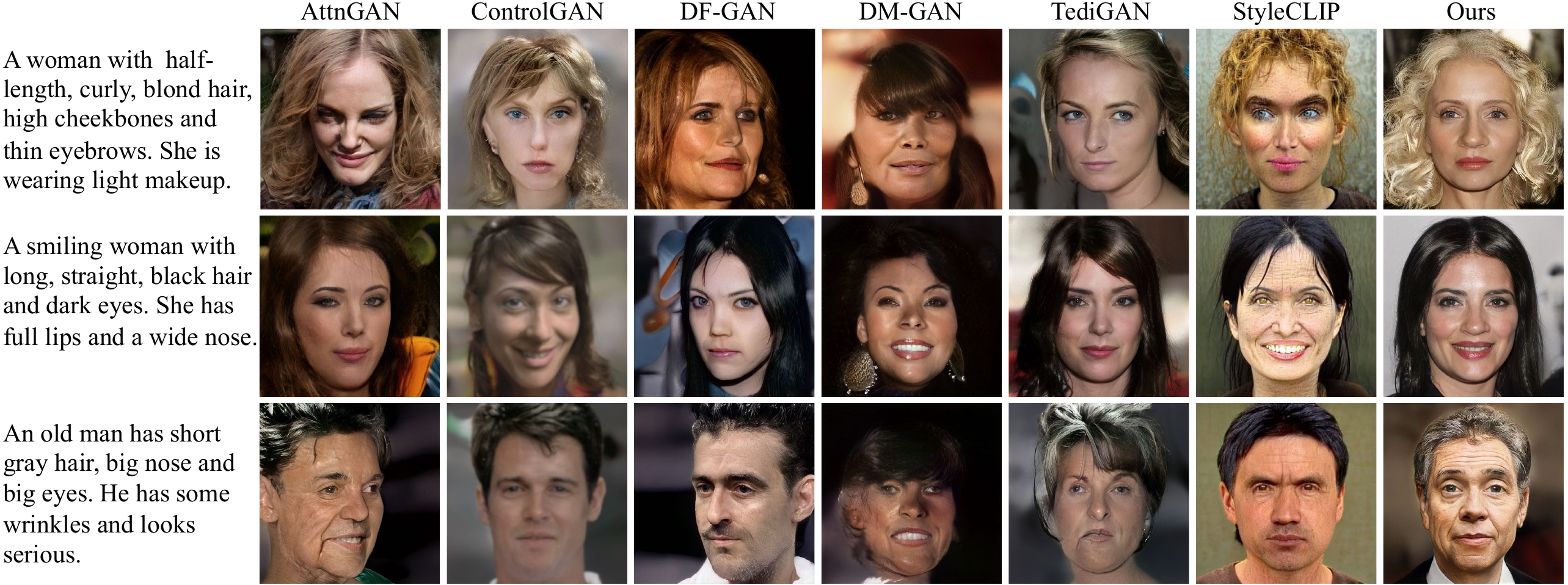}
    \vspace{-0.7cm}
    \caption{Qualitative comparison with existing methods, \emph{i.e.}, AttnGAN~\cite{xu2018attngan}, ControlGAN~\cite{li2019controllable}, DF-GAN~\cite{tao2022df}, DM-GAN~\cite{zhu2019dm}, TediGAN~\cite{xia2021tedigan}, and StyleCLIP~\cite{patashnik2021styleclip}. Our generated images using the text on the left show superior performance on fidelity and quality.}
    \label{fig:pic4}
    \vspace{-0.55cm}
\end{figure*}

\subsection{Text-guided Image Editing}
With the pre-trained CLIP2GAN model, we further apply the network to text-guided image editing applications.
Given a source image $x$, we are interested in editing certain regions of it by manipulating its latent codes $\mathbf{z}=\{\mathbf{z_i}\}_{i=1}^{i=18}$ to $\mathbf{z'}=\{\mathbf{z'_i}\}_{i=1}^{i=18}$ and getting a target image $x'$ that meets the editing requirements, which is expressed as follows,
\vspace{-0.15cm}
\begin{equation}
    \mathbf{z'_i} = \mathbf{z_i} + \beta \mathbf{n_i},
    \label{eq:inversion-1}
    \vspace{-0.1cm}
\end{equation}
\begin{equation}
    x' = G(\mathbf{z'}),
    \label{eq:inversion-2}
    \vspace{-0.1cm}
\end{equation}
where $\mathbf{n}=\{\mathbf{n_i}\}_{i=1}^{i=18}$ corresponds to the normal direction of a particular semantics of the latent space, and $\beta$ denotes the degree of editing semantics. 
That is, if the latent code moves in a certain direction, the semantics contained in the output image should vary accordingly. 
This requires our framework to locate the semantic direction $\mathbf{n}$ of the text and the latent code $\mathbf{z}$ of the image.

As shown in \cref{fig:pic3}, the text, $t$ \emph{i.e.}, simple descriptions on age, gender, hair, expression, \emph{etc.}, is fed into the CLIP~\cite{radford2021learning} text encoder to get CLIP text features.
Then the vector $\mathbf{n}$ in the StyleGAN latent space is derived by the pre-trained mapping network, which is considered as the normal direction of a particular semantics due to the mapping network bridging the CLIP feature space and the StyleGAN latent space.
Meanwhile, by putting the source image $x$ through the CLIP image encoder and the mapping network, the representation $\mathbf{z}$ of $x$ in the StyleGAN latent space is obtained.
They are formulated as follows,
\vspace{-0.15cm}
\begin{equation}
    \mathbf{n} = M[C_{text}(t)],
    \label{textediting}
    \vspace{-0.1cm}
\end{equation}
\begin{equation}
    \mathbf{z} = M[C_{img}(x)],
    \label{imgediting}
    \vspace{-0.1cm}
\end{equation}
where $C_{text}(\cdot)$ and $C_{img}(\cdot)$ denote the CLIP text and image encoder, respectively.

Finally, the semantic direction $\mathbf{n}$ and the latent feature $\mathbf{z}$ are weighted together and the weight of $\mathbf{n}$, \emph{i.e.}, $\beta$, ranges from 0 to 1, leading to the modified feature $\mathbf{z'}$ in the latent space of StyleGAN, and thus StyleGAN generates the corresponding images.
This allows the text description to control the degree of semantic modification of the image while not changing the identity features of the image itself. 
Our approach achieves high-quality text-guided image editing by simple arithmetic operations without optimization and additional network structures.

\vspace{-0.02cm}
\section{Experiments}
\label{sec:formatting}

\subsection{Experiments Setup}

\noindent \textbf{Datasets.}
To demonstrate the superiority of our method for text-guided face generation and face editing, we train on the CelebA-HQ~\cite{karras2018progressive} dataset and test using text descriptions from the Multi-modal CelebA-HQ (MM-CelebA-HQ)~\cite{xia2021tedigan} dataset.
The CelebA-HQ dataset is a high-quality version of the CelebA~\cite{liu2015deep} dataset, consisting of 30,000 images with a resolution of $1024^2$.
The MM-CelebA-HQ dataset creates 10 unique text descriptions for each image in CelebA-HQ.


\noindent \textbf{Evaluation metrics.}
We aim to assess visual quality, image accuracy, and realism for evaluation.
The visual quality of generated or manipulated images is evaluated through the widely-used Fréchet Inception Distance (FID)~\cite{heusel2017gans} metrics. 
FID measures the distance between two sets of images, computed by the mean value and covariance of the generated image set $(\mu_{Y},\Sigma_{Y})$ and the ground-truth image set $(\mu_{\hat{Y}},\Sigma_{\hat{Y}})$, which is formulated as follows,
\begin{equation}
    \text{FID}(Y,\hat{Y}) = \|\mu_{Y}-\mu_{\hat{Y}}\|^2_2 + \text{tr}(\Sigma_{Y}+\Sigma_{\hat{Y}}-2(\Sigma_{Y}\Sigma_{\hat{Y}})^{\frac{1}{2}}).
    \label{eq:fid}
\end{equation}
To evaluate the perceptual similarity between generated images and real images, we compute the average distance between them by the Learned Perceptual Image Patch Similarity (LPIPS)~\cite{zhang2018unreasonable} metrics, which is a weighted perceptual similarity between two images, computing on the features extracted from a pre-trained network.

In addition, accuracy and realism are evaluated through a user study. 
For image generation, the accuracy is evaluated by the similarity between the text and the generated image.
For image manipulation, accuracy is assessed by whether the visual properties of the modified image are aligned with the given description and whether contents unrelated to the text are preserved.
Realism is required to be judged as to which is more realistic and consistent with reality.
We tested accuracy and realism by collecting surveys on a random sample of 10 images from 20 people.

\subsection{Comparison with State-of-the-art Methods}
\begin{figure*}
    \centering
    \includegraphics[width=\linewidth]{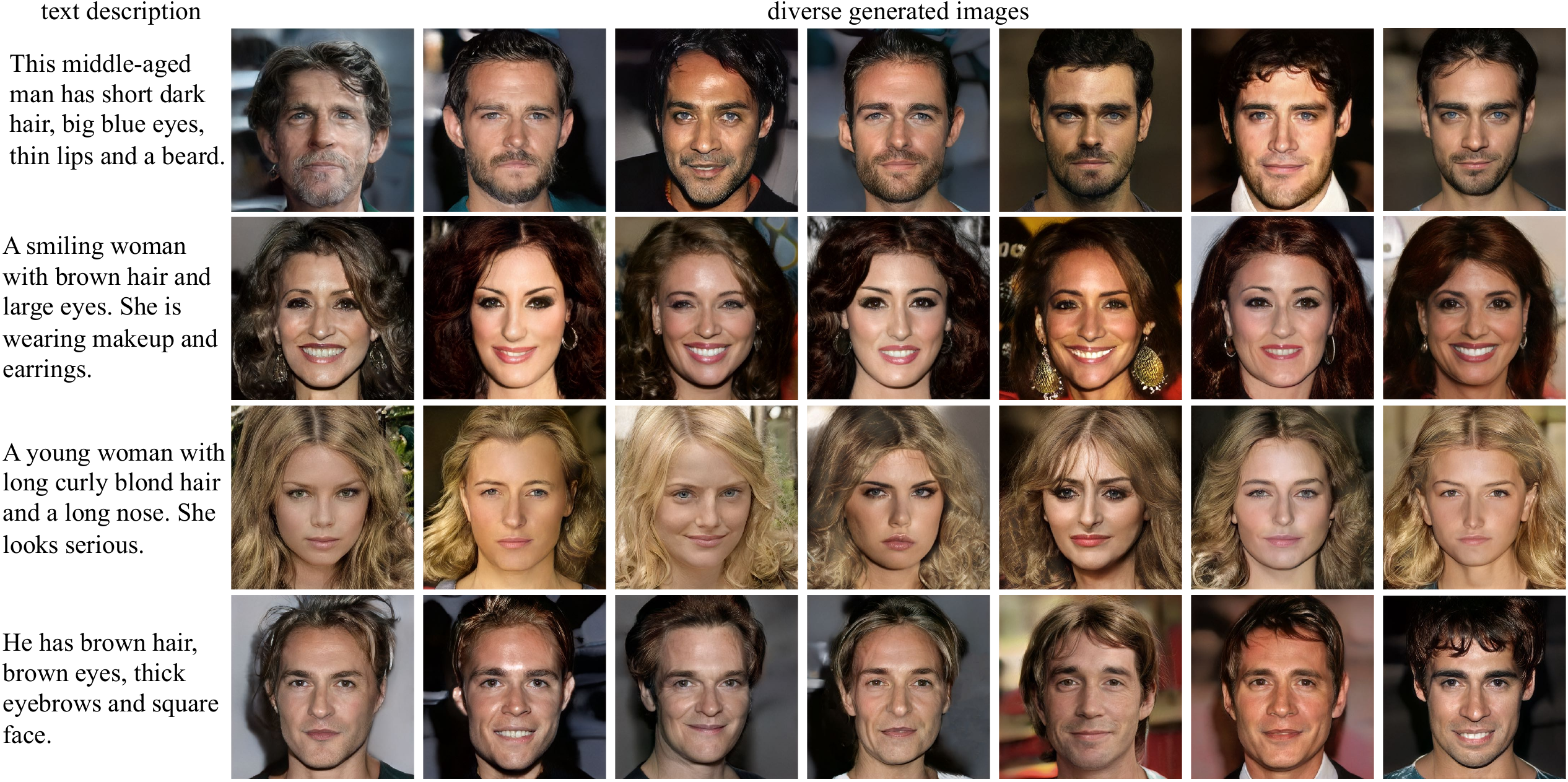}
    \vspace{-0.7cm}
    \caption{Diverse generation results from our framework. It is observed that our method can generate diverse results with high quality.}
    \label{fig:pic6}
    \vspace{-0.4cm}
\end{figure*}
We compare our method with several state-of-the-art methods of text-guided image generation, \emph{i.e.}, AttnGAN~\cite{xu2018attngan}, ControlGAN~\cite{li2019controllable}, DF-GAN~\cite{tao2022df}, DM-GAN~\cite{zhu2019dm}, TediGAN~\cite{xia2021tedigan}, and StyleCLIP~\cite{patashnik2021styleclip}.
We evaluate FID and LPIPS on a large number of samples generated from randomly selected text descriptions. 
Accuracy and realism are assessed by user studies. 
We generate 10 images using different methods and 20 users were asked to judge which one is the most realistic and aligned. 
The quantitative result is shown in \cref{tab:t2igeneration} and \cref{tab:t2iuserstudy}, where the other methods are trained and tested on the MM-CelebA-HQ dataset, while our method is trained on CelebA-HQ and tested with the help of the MM-CelebA-HQ text descriptions. 
From these tables, it is observed that our method achieves new state-of-the-art performance.

\begin{table}[t]
\footnotesize
\centering
\setlength{\tabcolsep}{9mm}{
\begin{tabular}{@{}l|cccc@{}}
\toprule
\textbf{Method} & \textbf{FID} & \textbf{LPIPS} \\ \midrule
AttnGAN~\cite{xu2018attngan}            & 125.98 & 0.512 \\
ControlGAN~\cite{li2019controllable}    & 116.32 & 0.522 \\
DF-GAN~\cite{tao2022df}                 & 137.60 & 0.581 \\
DM-GAN~\cite{zhu2019dm}                 & 131.05 & 0.544 \\
TediGAN~\cite{xia2021tedigan}           & 106.37 & 0.456 \\
StyleCLIP~\cite{patashnik2021styleclip} & 101.75 & 0.439 \\
\textbf{Ours} & \textbf{34.25} & \textbf{0.408} \\ \bottomrule
\end{tabular}
}
\vspace{-0.25cm}
\caption{Quantitative comparison with existing text-guided image generation methods on FID and LPIPS metrics. Our method outperforms previous algorithms on FID and LPIPS metrics.}
\label{tab:t2igeneration}
\vspace{-0.6cm}
\end{table}

Furthermore, we make a qualitative evaluation with several competitive methods, \emph{i.e.},  AttnGAN, ControlGAN, DF-GAN, DM-GAN, TediGAN, and StyleCLIP. 
From \cref{fig:pic4}, it is observed that our method has higher image quality and more realistic image results compared with previous methods. 
The images we generate correspond very closely to the text description as well as showing fine-grained details.
Benefiting from the CLIP model~\cite{radford2021learning}, we have an effective text-driven capability to generate face images with more text features in the category without being limited to text descriptions in the dataset.

\begin{figure}
    \centering
    \includegraphics[width=\linewidth]{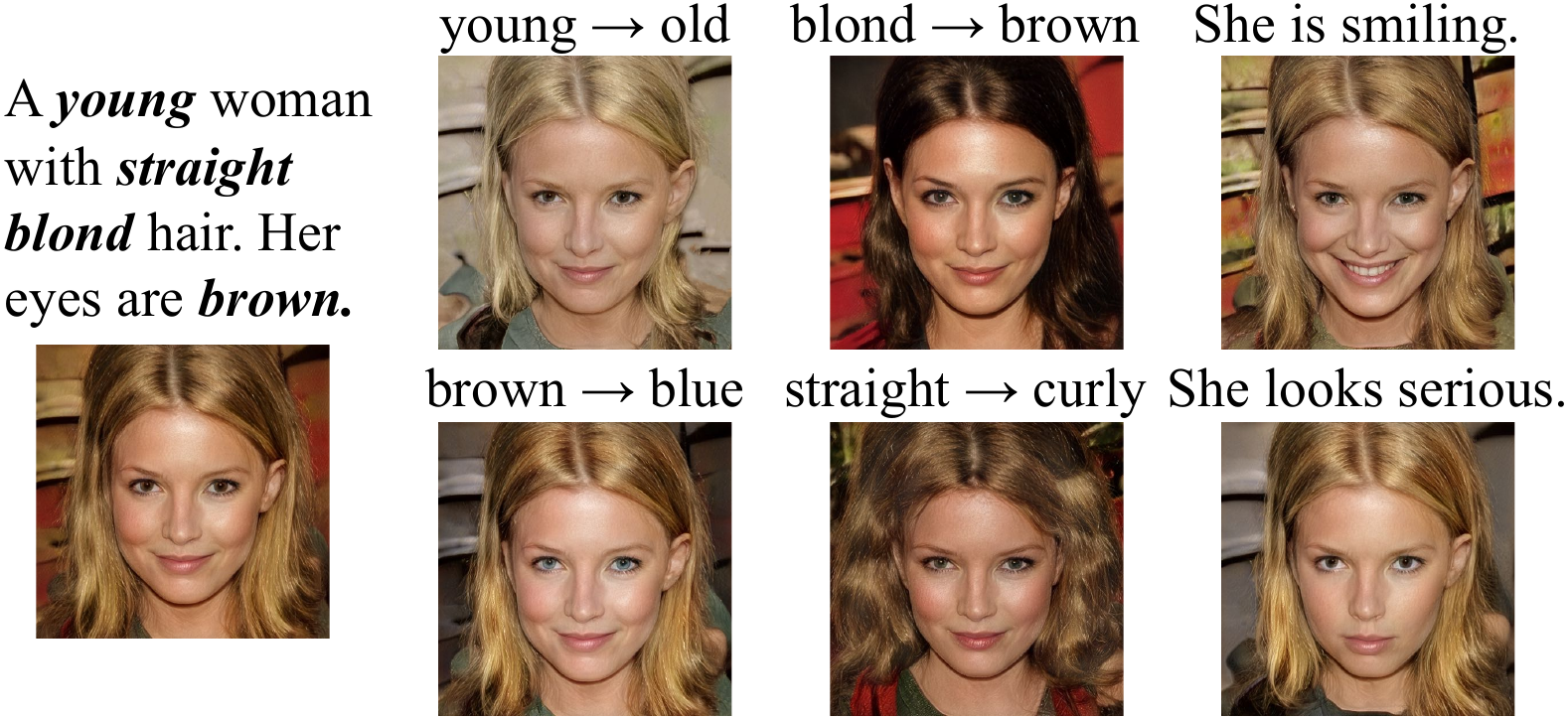}
    \vspace{-0.7cm}
    \caption{Generated result after modifying the text. The text on the top left is used as input to generate the image on the bottom left. By modifying the words marked in the text or adding descriptions of expressions as shown in the right half, the image generated from the newly input text only changes the corresponding features.}
    \label{fig:pic5}
    \vspace{-0.2cm}
\end{figure}

When some specific features in the text are changed, our model ensures that the image is modified only in the corresponding features, while other features, including identity features, are guaranteed to remain invariant.
This shows that our method is able to decouple different features with excellent robustness.
The text description and visual results are presented in \cref{fig:pic5}.


The other advantage is that our model can inherently generate diverse results given an arbitrary specific text description. 
With our approach, the generated images are guaranteed to be consistent with the given text features while other irrelevant features are varied to get multiple unique results. 
We present the text-guided image generation results in \cref{fig:pic6}, which demonstrates that our method can generate diverse results with high quality.

\begin{table}[t]
\footnotesize
\centering
\setlength{\tabcolsep}{5.8mm}{
\begin{tabular}{@{}l|cccc@{}}
\toprule
\textbf{Method} & \textbf{Acc.}\ (\%) & \textbf{Real.}\ (\%) \\ \midrule
Ours \emph{v.s.} AttnGAN~\cite{xu2018attngan}      & 83.0 & 84.5 \\
Ours \emph{v.s.} ControlGAN~\cite{li2019controllable}   & 80.5 & 79.0 \\
Ours \emph{v.s.} DF-GAN~\cite{tao2022df}       & 86.5 & 85.5 \\
Ours \emph{v.s.} DM-GAN~\cite{zhu2019dm}       & 88.0 & 91.5 \\
Ours \emph{v.s.} TediGAN~\cite{xia2021tedigan}      & 74.5 & 78.5 \\
Ours \emph{v.s.} StyleCLIP~\cite{patashnik2021styleclip}    & 70.5 & 95.5 \\ \bottomrule
\end{tabular}
}
\vspace{-0.25cm}
\caption{Paired user study between our method and existing text-guided image generation methods. Our method outperforms previous algorithms on Acc. and Real. metrics.}
\label{tab:t2iuserstudy}
\vspace{-0.5cm}
\end{table}

\begin{figure*}[t]
    \centering
    \includegraphics[width=\linewidth]{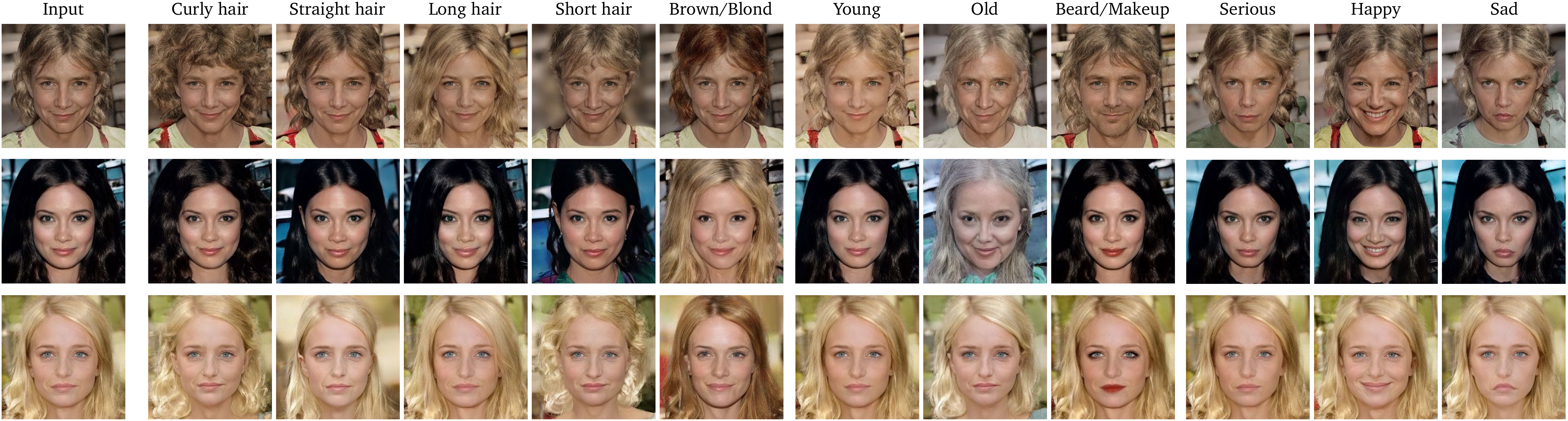}
    \vspace{-0.7cm}
    \caption{Visual results of text-guided image editing. Inputting the image on the left column and the text description above each of the other columns allows editing of the input image.}
    \label{fig:pic8}
    \vspace{-0.5cm}
\end{figure*}

\vspace{-0.1cm}
\subsection{Ablation Studies}
\vspace{-0.1cm}
There are several ablation experiments performed to demonstrate the effectiveness of the framework.
To evaluate the effectiveness and necessity of the network design, we modified the mapping network with a different number of network layers and the pre-trained models of CLIP.
From \cref{tab:ablation1}, it is observed that the framework performs worse in several metrics if the number of network layers is less. Whereas, with a higher number of layers, it is less significant for performance improvement, while greatly increasing the complexity of the network.
In addition, the feature space of CLIP has semantic meanings between images and text, thus using a stronger joint space (ViT/B-16) can improve the generated results.

\begin{table}[t]
\footnotesize
\centering
\setlength{\tabcolsep}{6.3mm}{
\begin{tabular}{@{}cc|ccc@{}}
\toprule
\multicolumn{2}{c|}{\textbf{Settings}} & \multicolumn{2}{c}{\textbf{Metrics}} \\ \midrule
\begin{tabular}[c]{@{}c@{}}Number \\ of Layers\end{tabular} & \begin{tabular}[c]{@{}c@{}}Pre-trained \\ CLIP~\cite{radford2021learning}\end{tabular} & FID & LPIPS \\ \midrule
4 & ViT/B-32 & 43.15 & 0.478 \\
8 & ViT/B-32 & 35.88 & 0.421 \\
12 & ViT/B-32 & 35.01 & 0.416 \\
4 & ViT/B-16 & 41.42 & 0.471 \\
8 & ViT/B-16 & 34.79 & 0.413 \\
\textbf{12} & \textbf{ViT/B-16} & \textbf{34.25} & \textbf{0.408} \\ \bottomrule
\end{tabular}
}
\vspace{-0.25cm}
\caption{Ablation of the network design, \emph{i.e.}, Number of Layers and Pre-trained CLIP, on CelebA-HQ~\cite{karras2018progressive} dataset. It is observed that using 12 layers of the mapping network and pre-trained CLIP-ViT/B-16 is the most appropriate of several settings.}
\label{tab:ablation1}
\vspace{-0.6cm}
\end{table}

We also investigate the impact of each component in the objective function. 
Ensuring that reconstruction loss, perceptual loss, and identity loss are preserved, we ablate by excluding adversarial loss and diversity loss one by one.
The results are shown in \cref{tab:ablation2}, and the performance decreases after each removal.
This is because removing the adversarial loss loses image realism while removing the diversity loss leads to pattern convergence.

\begin{table}[t]
\footnotesize
\centering
\setlength{\tabcolsep}{7.4mm}{
\begin{tabular}{@{}cc|ccc@{}}
\toprule
\multicolumn{2}{c|}{\textbf{Objective functions}} & \multicolumn{2}{c}{\textbf{Metrics}} \\ \midrule
$\mathcal{L}_{adv}$ & $\mathcal{L}_{div}$ & FID & LPIPS \\ \midrule
$\times$ & $\times$ & 55.48 & 0.589 \\
$\times$ & \checkmark & 52.76 & 0.561 \\
\checkmark & $\times$ & 38.44 & 0.454 \\
\checkmark & \checkmark & \textbf{34.25} & \textbf{0.408} \\ \bottomrule
\end{tabular}
}
\vspace{-0.25cm}
\caption{Ablation of the objective function, \emph{i.e.}, $\mathcal{L}_{adv}$ and $\mathcal{L}_{div}$, on CelebA-HQ~\cite{karras2018progressive} dataset, which denotes adversarial loss and diversity loss, respectively. Notably, our method achieves better performance with adversarial loss and diversity loss.}
\label{tab:ablation2}
\vspace{-0.6cm}
\end{table}

\subsection{Image Editing}
\vspace{-0.12cm}


To evaluate the performance of image editing, we use the FID metrics and conduct a user study. 
As shown in \cref{tab:t2iediting}, the results show that our method has better FID, accuracy, and realism than other methods~\cite{li2020manigan, xia2021tedigan, patashnik2021styleclip} on both CelebA-HQ dataset and arbitrary images that are not within the CelebA-HQ dataset. 
As well as generating high-quality text-guided images, we can achieve image editing for a given text description while ensuring that irrelevant content remains unchanged.

\begin{figure}[t]
    \centering
    \includegraphics[width=\linewidth]{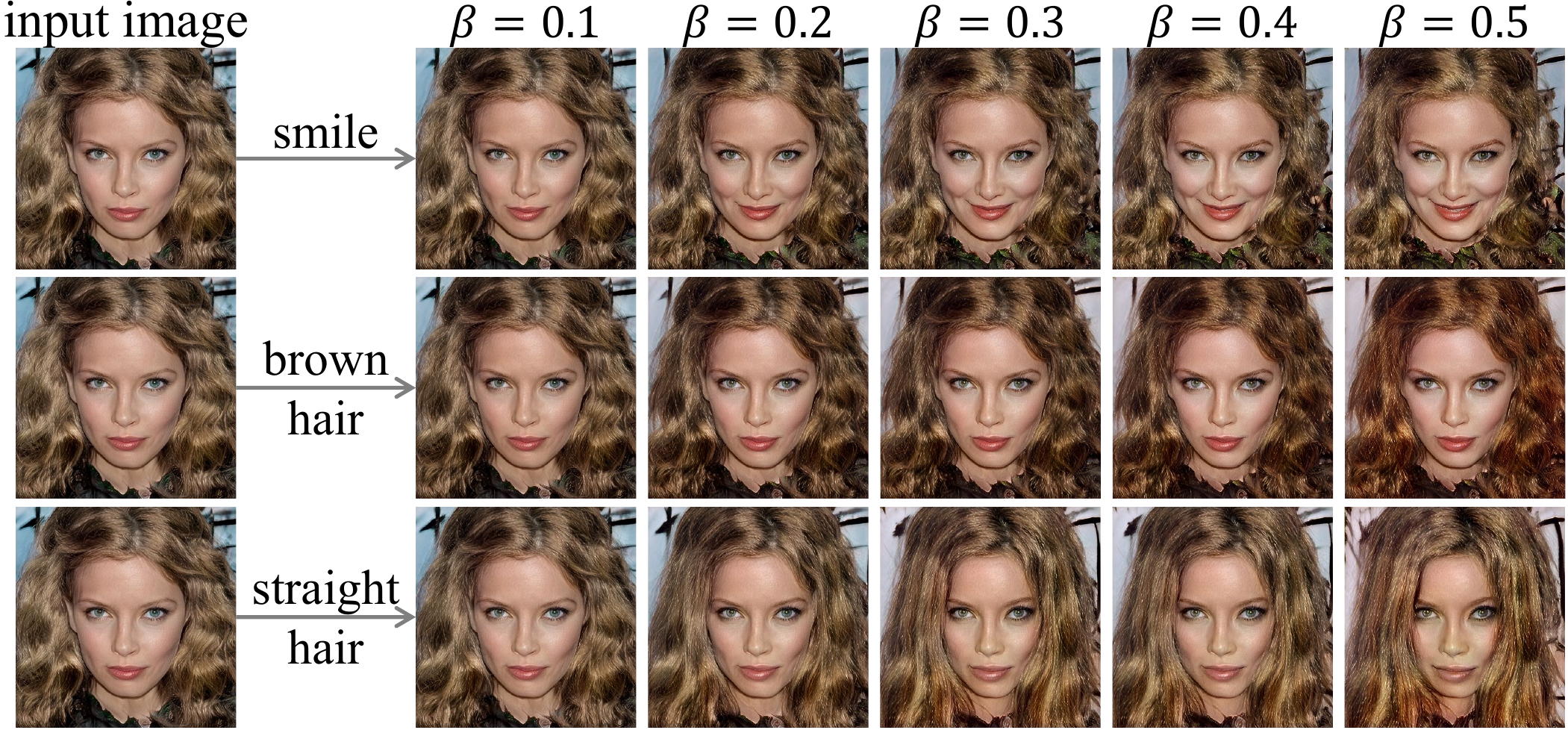}
    \vspace{-0.7cm}
    \caption{Varying degrees of text-guided image editing results. Notably, our method provides control on the modification degree.}
    \label{fig:pic7}
    \vspace{-0.25cm}
\end{figure}

\cref{fig:pic8} shows the visual results of our text-guided image editing.
With simple word descriptions, we can apply different manipulations to the image without changing other irrelevant features.
For the various text features used for modification, the extent to which the corresponding features of the image are changed is controlled by our framework, \emph{i.e.}, the value of the weight $\beta$, as illustrated in \cref{fig:pic7}.
Since CLIP is trained on a massive dataset of image-text alignment, which possesses a huge text latent space.
Therefore, using the pre-trained CLIP, our method can achieve text editing for numerous different features without being limited to some specific text descriptions.
\begin{table}[]
\footnotesize
\centering
\resizebox{\columnwidth}{!}{
\begin{tabular}{@{}l|ccc|ccc@{}}
\toprule
 & \multicolumn{3}{c|}{\textbf{CelebA-HQ}} & \multicolumn{3}{c}{\textbf{Not-in-CelebA-HQ}} \\ \cmidrule(l){2-7} 
\textbf{Method} & FID$\ \downarrow$ & Acc.$\ (\%)\uparrow$ & Real.$\ (\%)\uparrow$ & FID$\ \downarrow$ & Acc.$\ (\%)\uparrow$ & Real.$\ (\%)\uparrow$ \\ \midrule
ManiGAN~\cite{li2020manigan}            & 117.89 & 10.5 & 8.0  & 143.39 & 3.5  & 7.5  \\
TediGAN~\cite{xia2021tedigan}           & 107.25 & 16.0 & 18.5 & 135.47 & 18.0 & 24.5 \\
StyleCLIP~\cite{patashnik2021styleclip} & 86.82  & 28.5 & 21.5 & 106.24 &  31.5 & 21.5 \\
\textbf{Ours} & \textbf{35.66} & \textbf{45.0} & \textbf{52.0} & \textbf{48.14} & \textbf{47.0} & \textbf{46.5} \\ \bottomrule
\end{tabular}
}
\vspace{-0.35cm}
\caption{Quantitative comparison with existing methods on text-guided image editing. $\uparrow$ indicates the higher the better, while $\downarrow$ indicates the lower the better. our method achieves better performance on both the CelebA-HQ~\cite{karras2018progressive} dataset and arbitrary images outside the CelebA-HQ dataset.}
\label{tab:t2iediting}
\vspace{-0.6cm}
\end{table}
\vspace{-0.2cm}
\section{Conclusion}
\vspace{-0.12cm}
In this paper, we investigate vision-language models (CLIP) to generative models (StyleGAN) for the task of text-guided image generation and propose a novel framework named CLIP2GAN.
Specifically, the framework bridges the pre-trained CLIP and StyleGAN and implements training in a text-free way, where the trained model generates high-fidelity and high-quality images corresponding to the text description.
With the use of CLIP2GAN, we also achieve text-guided image manipulation that allows the editing of real images.
Extensive experiments demonstrate the effectiveness of our method.
Compared to previous methods, our framework achieves superior performance and shows better visual results.

{\small

\bibliographystyle{ieee_fullname}
\bibliography{PaperForReview}
}

\clearpage
This supplementary material provides additional implementation details and experimental results to support the main submission. First, we discuss the datasets used in the experiments in addition to those mentioned in the main paper and the detailed implementation setup. 
Next, we present additional ablation studies on using different architectures in our framework.
Finally, we show more qualitative and quantitative results of text-guided image generation and editing on different datasets.

\appendix

\section{Experiments Setup}
\label{sec:setup}
In this section, we provide more experiments setup on our proposed CLIP2GAN, including additional datasets used to demonstrate generalization capabilities and more implementation details.

\ 
\newline
\noindent \textbf{Datasets.}
In the main paper, we mention the CelebA-HQ~\cite{karras2018progressive} dataset for learning the image reconstruction and the Multi-Modal CelebA-HQ dataset~\cite{xia2021tedigan} for testing the effectiveness of the text-guided face image generation.
Besides, we have conducted experiments on AFHQ~\cite{choi2020stargan} dataset and LSUN~\cite{yu2015lsun} dataset.
The AFHQ dataset is an animal faces dataset consisting of 15,000 high-quality images at $512^2$ resolution, which includes three domains of cat, dog, and wildlife, each providing 5000 images.
We choose the cat and dog datasets of AFHQ for training.
The LSUN dataset contains around one million labeled images for each of the 10 scene categories and 20 object categories.
We choose the car and church datasets of LSUN for training.
For these two datasets, the identity loss of faces is removed.
The experiments on AFHQ and LSUN datasets also validate the superior performance of our method with the generality of its text-guided image generation function.

\
\newline
\noindent \textbf{Implementation details.}
The hyper-parameters in the framework are set as follows:
The trade-off parameter $\lambda_{rec}, \lambda_{LPIPS}, \lambda_{G}, \lambda_{id}, \lambda_{div}$ and $\lambda$ are set to 1, 1, 0.1, 1, 1 and 1, respectively, to ensure the training stability. 
For the whole framework, we utilize Adam optimizer~\cite{kingma2014adam}. 
The training lasts 100 epochs in total. 
The learning rate is set to $2\times10^{-3}$ and linearly reduces after 50 epochs. 
The whole framework is implemented by Pytorch and we perform experiments on NVIDIA RTX 3090.

\section{Ablation Studies}
\label{sec:ablation}

In this section, we provide additional ablation studies on using different pre-trained GANs in our framework and different locations for arithmetic operations during editing. By comparison, we demonstrate the effectiveness of our method to generate and edit with high quality and high fidelity.

\begin{table}[t]
\footnotesize
\centering
\resizebox{\columnwidth}{!}{
\begin{tabular}{@{}l|cc|cccc@{}}
\toprule
\multicolumn{1}{l|}{\multirow{2}{*}{\textbf{Model}}} & \multicolumn{2}{c|}{\textbf{Generation}}         & \multicolumn{3}{c}{\textbf{Editing}} \\ \cmidrule(l){2-6} 
\multicolumn{1}{c|}{}                                & FID                     & LPIPS                  & Operating Space   & FID     & LPIPS  \\ \midrule
\multirow{2}{*}{DCGAN~\cite{radford2015unsupervised}}                               & \multirow{2}{*}{183.37} & \multirow{2}{*}{0.562} & CLIP~\cite{radford2021learning}              & n/a       & n/a      \\
                                                     &                         &                        & GAN               & 196.76  & 0.583  \\
\multirow{2}{*}{\textbf{StyleGAN2}~\cite{karras2020analyzing}}                           & \multirow{2}{*}{\textbf{34.25}}  & \multirow{2}{*}{\textbf{0.408}} & CLIP~\cite{radford2021learning}              & 60.54   & 0.505  \\
                                                     &                         &                        & \textbf{GAN}               & \textbf{35.66}   & \textbf{0.417}  \\ \bottomrule
\end{tabular}
}
\caption{Additional ablation studies of different architectures on CelebA-HQ~\cite{karras2018progressive} dataset. The results show that our method works on other GANs as well, and the editing performance of the operation in the latent space of the GAN is better than that in the CLIP feature space.}
\label{tab:abla}
\end{table}

\
\newline
\noindent \textbf{Effect of different pre-trained GANs.}
To demonstrate that the mapping network in CLIP2GAN can map the feature space of CLIP~\cite{radford2021learning} to the latent space of any GAN, we replace StyleGAN~\cite{karras2019style} in the framework with DCGAN~\cite{radford2015unsupervised} and perform learning.
The capability of DCGAN is weak, which leads to the generated images looking blurry and of poor quality.
Fortunately, the text-guided generated images are still able to satisfy the requirements of the attributes mentioned in the text, which indicates that our model is effective for the latent space of any GAN.
We present quantitative analysis in \cref{tab:abla}.

\
\newline
\noindent \textbf{Effect of the editing operation location.}
For image editing, we choose to perform arithmetic operations on text features and image features at different locations, \emph{i.e.}, the feature space of CLIP, and the latent space of StyleGAN.
As shown in \cref{tab:abla}, better editing results can be obtained by performing arithmetic operations in the StyleGAN space.
The latent space of StyleGAN possesses better disentanglement properties than the feature space of CLIP, in which it is easier to attach text features to image features to achieve the generation of high-quality edited images.

\section{Results of Generation and Editing}
\label{sec:results}

In this section, We first show more generation and editing results to further demonstrate the effectiveness of our method. Then we present additional results on different datasets, which show the generalization capability of our method. 
With all of the above, we can produce high-quality and high-fidelity images.

\subsection{Additional Qualitative Results}
\label{subsec:add}
\begin{figure*}[t]
    \centering
    \includegraphics[width=\linewidth]{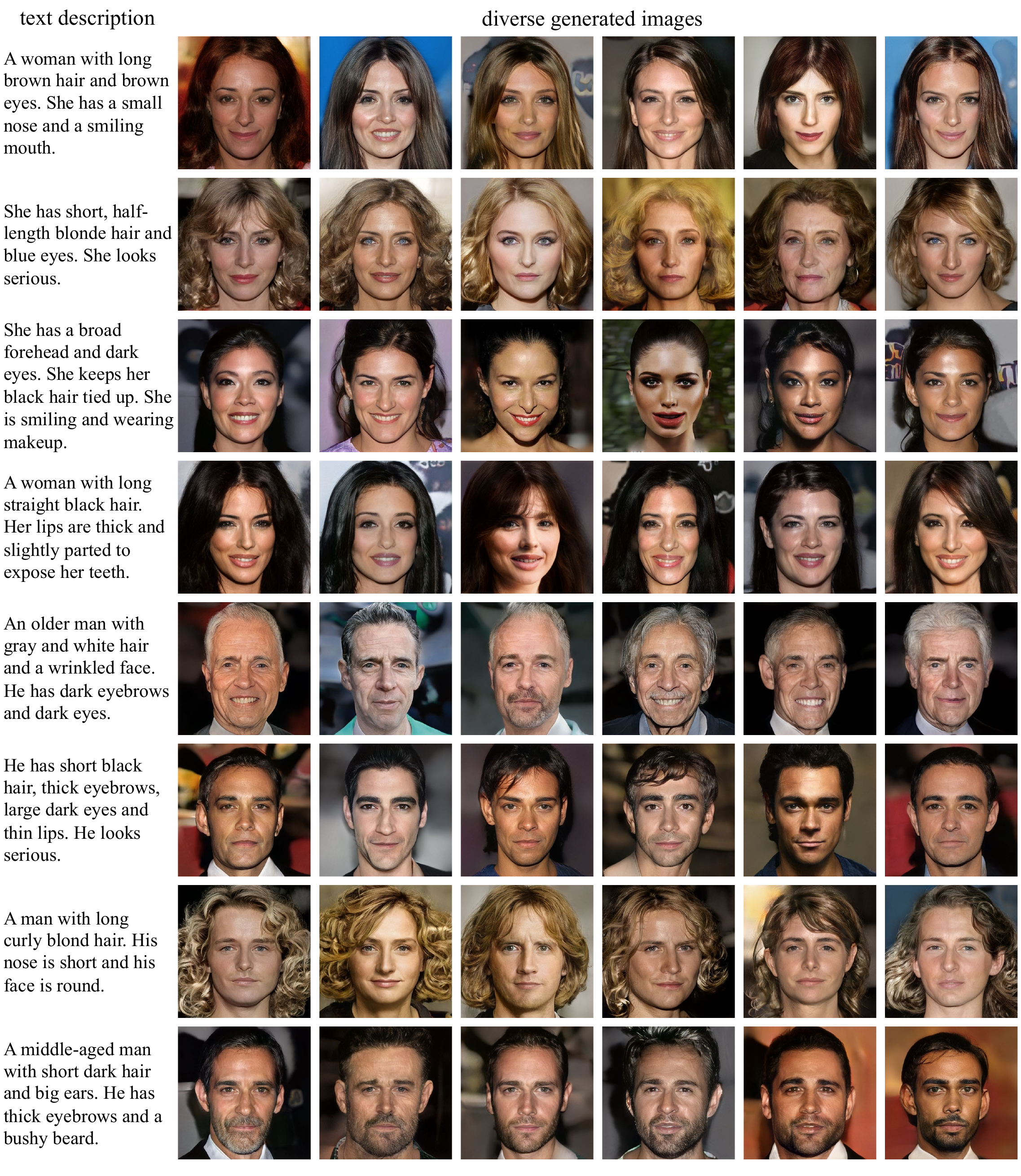}
    \caption{Additional qualitative results of text-guided image generation on CelebA-HQ~\cite{karras2018progressive} dataset. Given an arbitrary text on the left, we are able to generate diverse images on the right.}
    \label{fig:supp_pic1}
\end{figure*}

\begin{figure*}[t]
    \centering
    \includegraphics[width=\linewidth]{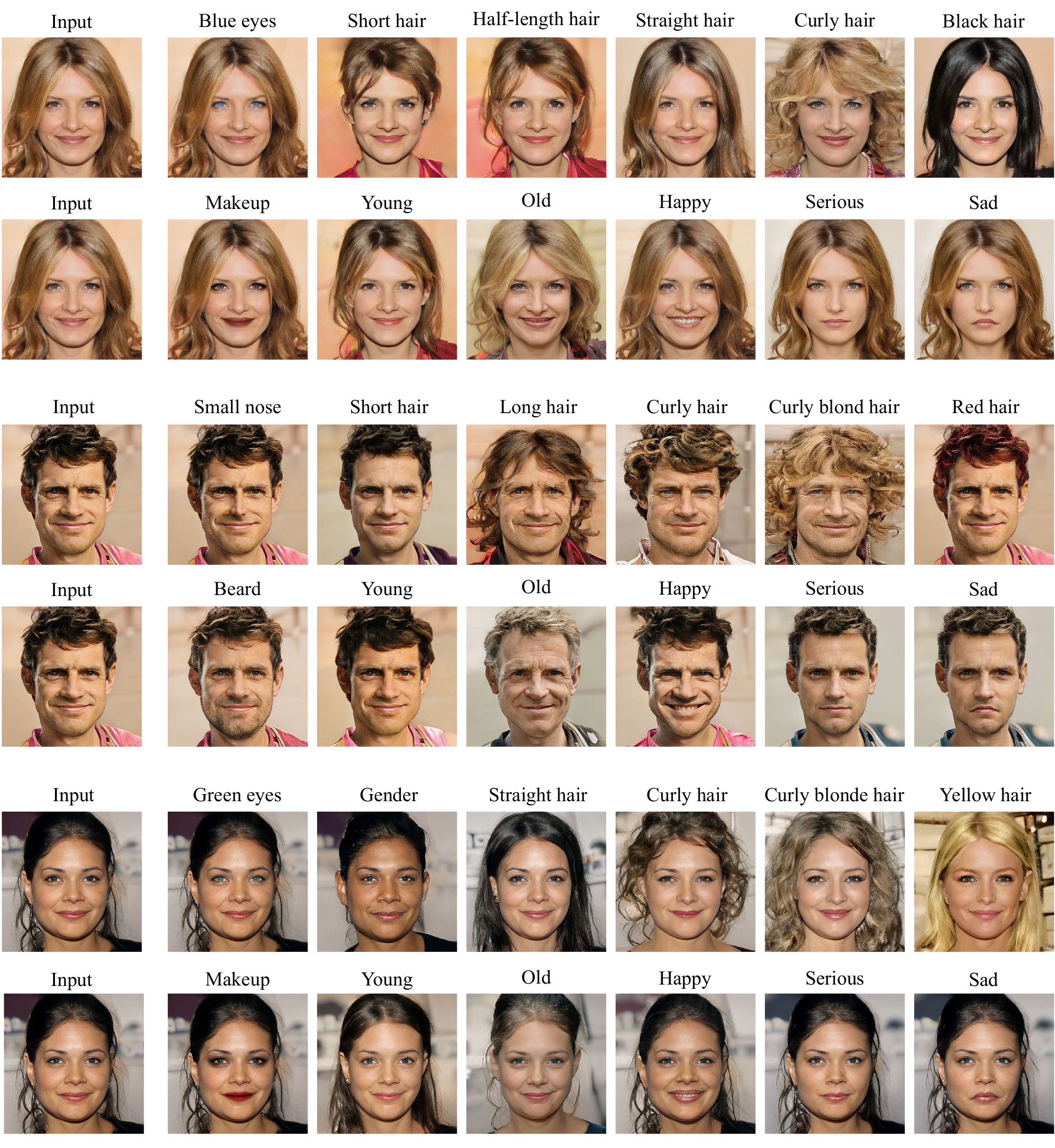}
    \caption{Additional qualitative results of text-guided image editing on CelebA-HQ~\cite{karras2018progressive} dataset. Given an arbitrary image on the left, we are able to edit the image according to the different textual descriptions provided above, without changing other irrelevant attributes, which are shown on the right.}
    \label{fig:supp_pic2}
\end{figure*}

We show more qualitative results of text-guided face generation and editing. As shown in \cref{fig:supp_pic1}, our method is able to generate diverse face images with high fidelity and high quality with an arbitrary text input. Also, as shown in \cref{fig:supp_pic2}, given a real face image and a textual description, we can edit the image according to text attributes without changing other irrelevant attributes.

\subsection{Results on Different Datasets}
\label{subsec:datasets}
In addition to face generation and editing, we also conduct experiments on other datasets, \emph{i.e.}, the cat and dog datasets of AFHQ~\cite{choi2020stargan}, the church datasets of LSUN~\cite{yu2015lsun}. 
Due to the lack of corresponding text descriptions for these datasets mentioned above, it is difficult for us to compare them with other methods.
The quantitative results of our method on these three datasets are shown in \cref{tab:3dataset}.
Evidently, the text-guided image generation model still works well on the other datasets, which indicates the generality of our model on various image datasets that do not require text descriptions.
\begin{table}[t]
\footnotesize
\centering
\setlength{\tabcolsep}{8.1mm}{
\begin{tabular}{@{}l|cccc@{}}
\toprule
\textbf{Dataset} & \textbf{FID} & \textbf{LPIPS} \\ \midrule
CelebA-HQ~\cite{karras2018progressive} & 34.25 & 0.408 \\
CAT of AFHQ~\cite{choi2020stargan}     & 62.33 & 0.491 \\
DOG of AFHQ~\cite{choi2020stargan}     & 68.17 & 0.487 \\
CHURCH of LSUN~\cite{yu2015lsun}       & 92.84 & 0.534 \\ \bottomrule
\end{tabular}
}
\caption{Quantitative results on different datasets, \emph{i.e.}, CelebA-HQ dataset, the cat and dog datasets of AFHQ, the church datasets of LSUN. In each dataset, our method achieves excellent results for text-guided image generation.}
\label{tab:3dataset}
\end{table}

\begin{figure*}[t]
    \centering
    \vspace{-1.2cm}
    \includegraphics[width=\linewidth]{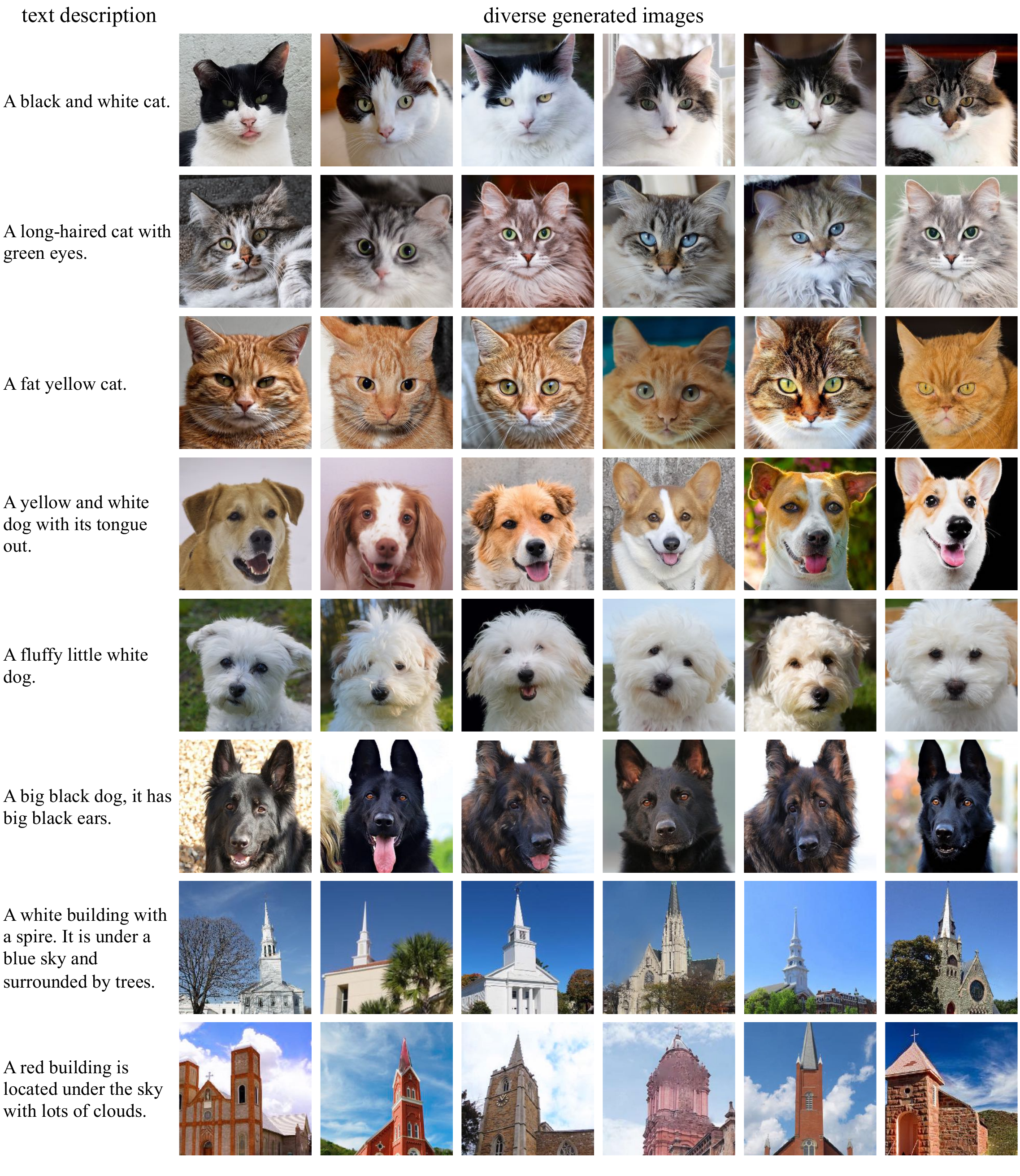}
    \caption{Qualitative results of text-guided image generation on CAT of AFHQ~\cite{choi2020stargan} dataset, DOG of AFHQ~\cite{choi2020stargan} dataset, and CHURCH of LSUN~\cite{yu2015lsun} dataset. Given an arbitrary text on the left, we are able to generate diverse images on the right.}
    \label{fig:supp_pic3}
\end{figure*}

\begin{figure*}[t]
    \centering
    \includegraphics[width=0.7\linewidth]{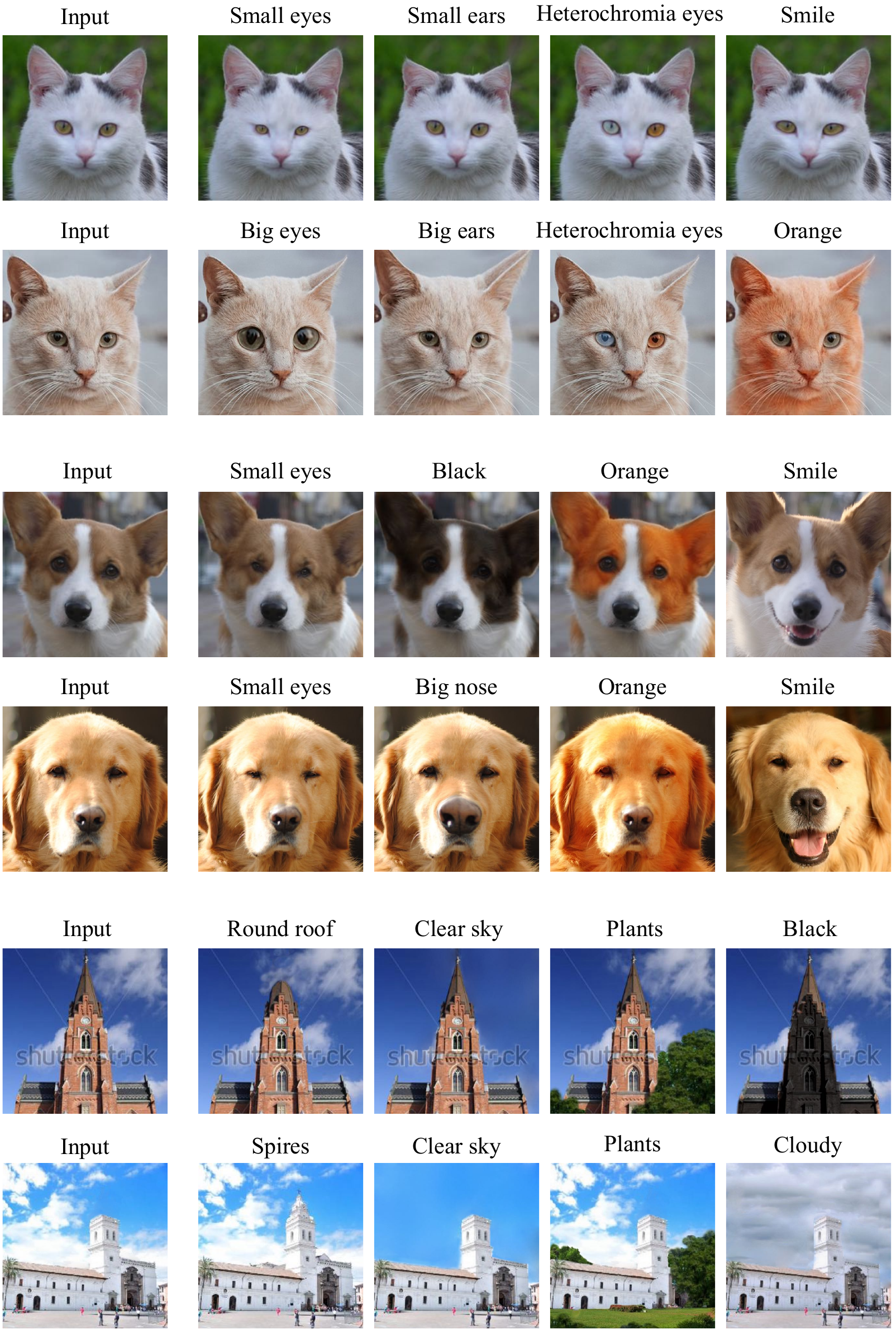}
    \vspace{0.5cm}
    \caption{Qualitative results of text-guided image editing on CAT of AFHQ~\cite{choi2020stargan} dataset, DOG of AFHQ~\cite{choi2020stargan} dataset, and CHURCH of LSUN~\cite{yu2015lsun} dataset. Given an arbitrary image on the left, we are able to edit the image according to the different textual descriptions provided above, without changing other irrelevant attributes, which are shown on the right.}
    \label{fig:supp_pic4}
\end{figure*}

We also present the qualitative results of these three datasets. \cref{fig:supp_pic3} shows the results of the text-guided image generation on different datasets. Besides, \cref{fig:supp_pic4} shows the results of text-guided image editing on different datasets. The generated and edited results we achieved are diverse, high fidelity, and high quality.

\end{document}